\documentclass[10pt,twocolumn,letterpaper]{article}

\usepackage{cvpr}
\usepackage{times}
\usepackage{epsfig}
\usepackage{graphicx}
\usepackage{amsmath}
\usepackage{amssymb}

\usepackage[breaklinks=true,bookmarks=false]{hyperref}

\cvprfinalcopy 

\def\cvprPaperID{2662} 

\ifcvprfinal\fi
\begin{document}

\title{Reconstruction Network for Video Captioning}

\author{Bairui Wang$^\ddagger$ \qquad Lin Ma$^{\dagger}$\thanks{Corresponding authors} \qquad Wei Zhang$^{\ddagger\ast}$ \qquad Wei Liu$^\dagger$  \\
$^\dagger$Tencent AI Lab\qquad $^\ddagger$School of Control Science and Engineering, Shandong University\\
{\tt\small \{bairuiwong,forest.linma\}@gmail.com\qquad davidzhang@sdu.edu.cn\qquad wliu@ee.columbia.edu}
}
\maketitle
\thispagestyle{empty}

\begin{abstract}
In this paper, the problem of describing visual contents of a video sequence with natural language is addressed. Unlike previous video captioning work mainly exploiting the cues of video contents to make a language description, we propose a reconstruction network (RecNet) with a novel encoder-decoder-reconstructor architecture, which leverages both the forward (video to sentence) and backward (sentence to video) flows for video captioning. Specifically, the encoder-decoder makes use of the forward flow to produce the sentence description based on the encoded video semantic features. Two types of reconstructors are customized to employ the backward flow and reproduce the video features based on the hidden state sequence generated by the decoder. The generation loss yielded by the encoder-decoder and the reconstruction loss introduced by the reconstructor are jointly drawn into training the proposed RecNet in an end-to-end fashion. Experimental results on benchmark datasets demonstrate that the proposed reconstructor can boost the encoder-decoder models and leads to significant gains in video caption accuracy.
\end{abstract}

\section{Introduction}
\label{sec:intro}
Describing visual contents with natural language automatically has received increasing attention in both the computer vision and natural language processing communities. It can be applied in various practical applications, such as image and video retrieval \cite{song2017quantization,wang2017survey,ma2015multimodal},
answering questions from images~\cite{ma2016learning},
and assisting people who suffer from vision disorders~\cite{Voykinska2016How}.

Previous work predominantly focused on describing still images with natural language \cite{karpathy2014deep,Vinyals_2015_CVPR,vinyals2017show,ren2017deep,jiang2018learning,chen2016sca}. Recently, researchers have strived to generate sentences to describe video contents \cite{xu2015jointly,donahue2015long,DBLP:conf/iccv/VenugopalanRDMD15,venugopalan2015translating,pan2016video}. Compared to image captioning, describing videos is more challenging as the amount of information (\textit{e.g.}, objects, scenes, actions, \etc.) contained in videos is much  more sophisticated than that in still images. More importantly, the temporal dynamics within video sequences need to be adequately captured for captioning, besides the spatial content modeling.

\begin{figure}[t]
  \centering
  \includegraphics[width=1\linewidth]{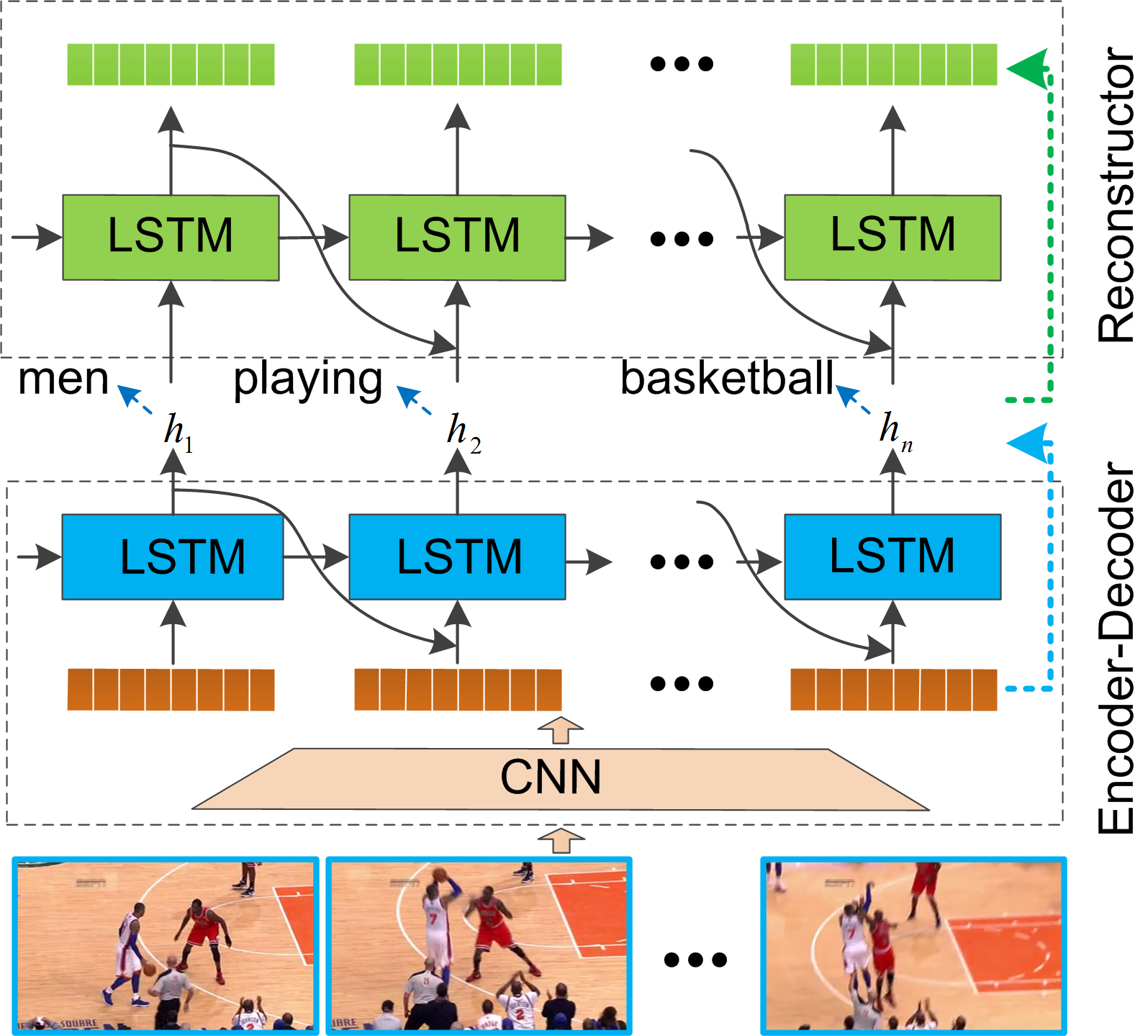}
  \caption{The proposed RecNet with an encoder-decoder-reconstructor architecture. The encoder-decoder relies on the forward flow from video to caption (blue dotted arrow), in which the decoder generates caption with the frame features yielded by the encoder. The reconstructor exploiting the backward  flow from caption to video (green dotted arrow), takes the hidden state sequence of the decoder as input and reproduces the visual features of the video.} 
  \label{fig1}
\end{figure}

Recently, the encoder-decoder architecture, has been widely adopted for video captioning \cite{donahue2015long,ramanishka2016multimodal,jin2016describing,DBLP:conf/iccv/YaoTCBPLC15,gao2017video,song2017hierarchical,pan2016jointly,pan2016video,liu2017video}, as shown in Fig.~\ref{fig1}. However, the encoder-decoder architecture only relies on the forward flow (video to sentence), but does not consider the information from sentence to video, named as backward flow. Usually the encoder is a convolutional neural network (CNN) capturing the image structure to yield its semantic representation. For a given video sequence, the yielded semantic representations by the CNN are further fused together to exploit the video temporal dynamics and generate the video representation. The decoder is usually a long short-term memory (LSTM)~\cite{hochreiter1997long} or a gated recurrent unit (GRU)~\cite{cho2014properties}, which is popular in processing sequential data \cite{zhang2017learning}.
LSTM and GRU generate the sentence fragments one by one, and ensemble them to form one sentence. The semantic information from target sentences to source videos are never included.
Actually, the backward flow can be yielded by the dual learning mechanism that has been introduced into neural machine translation (NMT)~\cite{tu2017neural,he2016dual} and image segmentation~\cite{luo2017deep}. This mechanism reconstructs source from target when the target is achieved and demonstrates that backward flow from target to source improves performance.

To well exploit the backward flow, we refer to idea of dual learning and propose an encoder-decoder-reconstructor architecture shown in Fig.~\ref{fig1}, dubbed as RecNet, to address video captioning. Specifically, the encoder-decoder yields the semantic representation of each video frame and subsequently generates a sentence description. Relying on the backward flow, the reconstructor, realized by LSTMs, aims at reproducing the original video feature sequence based on the hidden state sequence of the decoder. The reconstructor, targeting at minimizing the differences between original and reproduced video features, is expected to further bridge the semantic gap between the natural language captions and video contents.

To summarize, the contributions of this work lie in three-fold.
\begin{itemize}
\item We propose a novel reconstruction network (RecNet) with an encoder-decoder-reconstructor architecture to exploit both the forward (video to sentence) and backward (sentence to video) flows for video captioning.
\item Two types of reconstructors are customized to restore the video global and local structures, respectively.
\item Extensive results on benchmark datasets indicate that the backward flow is well addressed by the proposed reconstructor and significant gains on video captioning are achieved.
\end{itemize}

\section{Related Work}
In this section, we first introduce two types of video captioning: template-based approaches \cite{kojima2002natural,guadarrama2013youtube2text,rohrbach2013translating,rohrbach2014coherent,xu2015jointly} and sequence learning approaches \cite{DBLP:conf/iccv/YaoTCBPLC15,DBLP:conf/iccv/VenugopalanRDMD15,venugopalan2015translating,donahue2015long,ramanishka2016multimodal,jin2016describing,zhang2016automatic,pan2016jointly,pan2016video,shen2017weakly,liu2017video}, then introduce the application of dual learning.

\subsection{Template-based Approaches}
Template-based methods first define some specific rules for language grammar, and then parse the sentence into several components such as subject, verb, and object. The obtained sentence fragments are associated with words detected from the visual content to produce the final description about the input video with predefined templates. For example, a concept hierarchy of actions was introduced to describe human activities in \cite{kojima2002natural}, while a semantic hierarchy was defined in \cite{guadarrama2013youtube2text} to learn the semantic relationship between different sentence fragments. In \cite{rohrbach2013translating}, the conditional random field (CRF) was adopted to model the connections between objects and activities of the visual input and generate the semantic features for description. Besides, Xu \etal proposed a unified framework consisting of a semantic language model, a deep video model, and a joint embedding model to learn the association between videos and natural sentences \cite{xu2015jointly}. However, as stated in \cite{pan2016video}, the aforementioned approaches highly depend on the predefined template and are thus limited by the fixed syntactical structure, which is inflexible for sentence generation.

\begin{figure*}
  \centering
  \includegraphics[width=1\linewidth]{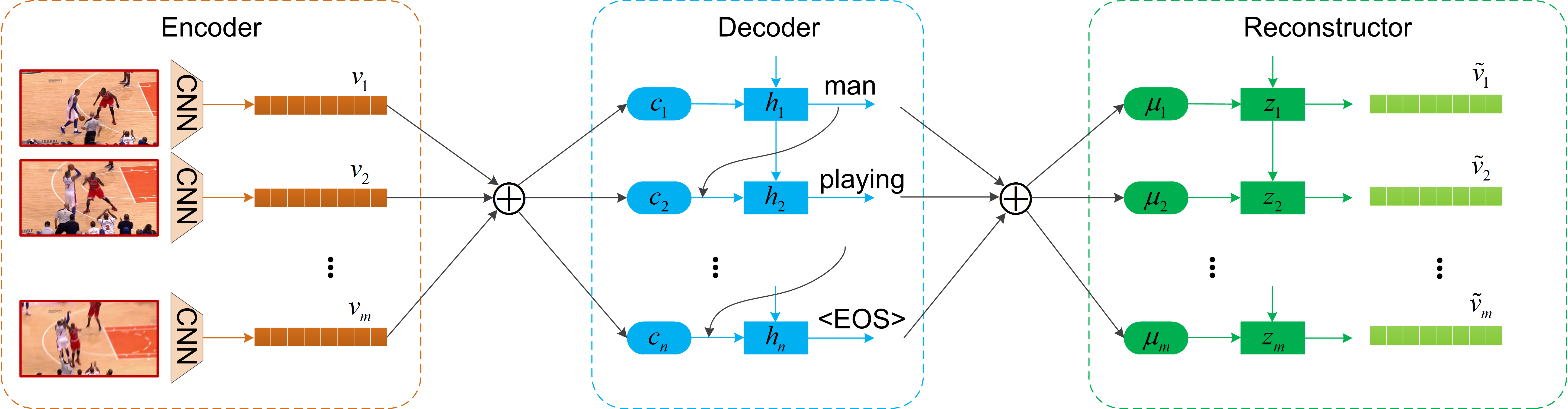}
  \caption{The proposed RecNet consists of three parts: the CNN-based encoder which extracts the semantic representations of the video frames, the LSTM-based decoder which generates natural language for visual content description, and the reconstructor which exploits the backward flow from caption to visual contents to reproduce the frame representations.} 
  \label{fig2}
\end{figure*}

\subsection{Sequence Learning Approaches}
Compared with the template-based methods, the sequence learning approaches aim to directly produce the sentence description about the visual input with more flexible syntactical structures. For example, in \cite{venugopalan2015translating}, the video representation was obtained by averaging each frame feature extracted by a CNN, and then fed to LSTMs for sentence generation. In \cite{pan2016jointly}, the relevance between video context and sentence semantics was considered as a regularizer in the LSTM. However, since simple mean pooling is used, the temporal dynamics of the video sequence are not adequately  addressed. Yao \etal. introduced an attention mechanism to assign weights to the features of each frame and then fused them based on the attentive weights \cite{DBLP:conf/iccv/YaoTCBPLC15}. Venugopalan \etal. proposed S2VT \cite{DBLP:conf/iccv/VenugopalanRDMD15}, which included the temporal information with optical flow and employed LSTMs in both the encoder and decoder. To exploit both temporal and spatial information, Zhang and Tian proposed a two-stream encoder comprised of two 3D CNNs \cite{DBLP:journals/corr/TranBFTP14,DBLP:conf/cvpr/KarpathyTSLSF14} and one parallel fully connected layer to learn the features from the frames \cite{zhang2016automatic}. Besides, Pan \etal. proposed a transfer unit to model the high-level semantic attributes from both images and videos, which are rendered as the complementary knowledge to video representations for boosting sentence generation \cite{pan2016video}.

In this paper, our proposed RecNet can be regarded as a sequence learning method. However, unlike the above conventional encoder-decoder models which only depend on the forward flow from video to sentence, RecNet can also benefit the backward flow from sentence to video. By fully considering the bidirectional flows between video and sentence, RecNet is capable of further boosting the video captioning.

\subsection{Dual Learning Approaches}
As far as we know, dual learning mechanism has not been employed in video captioning but widely used in NMT~\cite{tu2017neural,he2016dual,xia2017dual}. In~\cite{tu2017neural}, the source sentences are reproduced from the target side hidden states, and the accuracy of reconstructed source provides a constraint for decoder to embed more information of source language into target language. In~\cite{he2016dual}, the dual learning is employed to train model of inter-translation of English-French, and get significantly improvement on tasks of English to French and French to English.

\section{Architecture}
We propose a novel RecNet with an encoder-decoder-reconstructor architecture for video captioning, which works in an end-to-end manner. The reconstructor imposes one constraint that the semantic information of one source video can be reconstructed from the hidden state sequence of the decoder. The encoder and decoder are thus encouraged to embed more semantic information about the source video. As illustrated in Fig.~\ref{fig2}, the proposed RecNet consists of three components, specifically the encoder, the decoder, and the reconstructor.
Moreover, our designed reconstructor can collaborate with different classical encoder-decoder architectures for video captioning. 
In this paper, we employ the attention-based video captioning \cite{DBLP:conf/iccv/YaoTCBPLC15} and S2VT \cite{DBLP:conf/iccv/VenugopalanRDMD15}. We first briefly introduce the encoder-decoder model for video captioning. Afterwards, the proposed reconstructors with two different architectures are described.

\subsection{Encoder-Decoder}
\label{ed}
The aim of video captioning is to generate one sentence $\mathbf{S}=\{\mathbf{s}_1,\mathbf{s}_2, \dots ,\mathbf{s}_n\}$ to describe the content of one given video $\mathbf{V}$. Classical encoder-decoder architectures directly model the captioning generation probability word by word:
\begin{equation}
P( \mathbf{S}|\mathbf{V} )=\prod_{i=1}^{n}  P\left ( \mathbf{s}_{i} | \mathbf{s}_{< i},\mathbf{V}; \theta \right ),
\end{equation}
where $\theta$ keeps the parameters of the encoder-decoder model. $n$ denotes the length of the sentence, and $\mathbf{s}_{< i}$ (\textit{i.e.}, $\{ \mathbf{s}_1, \mathbf{s}_2, \dots, \mathbf{s}_{i-1} \}$) denotes the generated partial caption.

\vspace{5pt}
\noindent\textbf{Encoder.} {To generate reliable captions, visual features need to be extracted to capture the high-level semantic information about the video.
Previous methods usually rely on CNNs, such as AlexNet \cite{venugopalan2015translating}, GoogleNet \cite{DBLP:conf/iccv/YaoTCBPLC15}, and VGG19 \cite{xu2016msr} to
encode each video frame into a fixed-length representation with the high-level semantic information. By contrast, in this work, considering a deeper network is more plausible for feature extraction, we advocate using Inception-V4 \cite{DBLP:conf/aaai/SzegedyIVA17} as the encoder. In this way, the given video sequence is encoded as a sequential representation $\mathbf{V}=\{\mathbf{v}_1, \mathbf{v}_2, \dots, \mathbf{v}_m\}$, where $m$ denotes the total number of the video frames.}

\vspace{5pt}
\noindent\textbf{Decoder.} Decoder aims to generate the caption word by word based on the video representation.
LSTM with the capabilities of modeling long-term temporal dependencies are used to decode video representation to video captions word by word. To further exploit the global temporal information of videos, a temporal attention mechanism \cite{DBLP:conf/iccv/YaoTCBPLC15} is employed to encourage the decoder to selecting the key frames/elements for captioning.

During the captioning process, the $i_{th}$ word prediction is generally made by LSTM:
\begin{equation}
P\left ( \mathbf{s}_{i} | \mathbf{s}_{< i},\mathbf{V}, \theta \right ) \propto \mathbf{exp}\big(f(\mathbf{s}_{i-1},h_i,c_i;\theta)\big),
\end{equation}
where $f$ represents the LSTM activation function, $h_i$ is the $i_{th}$ hidden state computed in the LSTM, and $c_i$ denotes the $i_{th}$ context vector computed with the temporal attention mechanism. The temporal attention mechanism is used to assign weight $\alpha_{i}^{t}$ to the representation of each frame $\left \{ \mathbf{v}_{1},\mathbf{v}_{2}, \dots,\mathbf{v}_{m} \right \}$ at the time step $t$ as follows:
\begin{equation}
\label{eq:3}
c_t=\sum_{i=1}^{m}\alpha_{i}^{t}\mathbf{v}_{i},
\end{equation}
where $m$ denotes the number of the video frames. With the $(i-1)_{th}$ hidden state $h_{i-1}$ summarizing all the current generated words, the attention weight $\alpha_{i}^t$ reflects the relevance of the $i_{th}$ temporal feature in the video sequence given all the previously generated words. As such, the temporal attention strategy allows the decoder to select a subset of key frames to generate the word at each time step, which can improve the video captioning performance as demonstrated in \cite{DBLP:conf/iccv/YaoTCBPLC15}.

The encoder-decoder model can be jointly trained by minimizing the {negative log likelihood} to produce the correct description sentence given the video as follows:
\begin{equation}
\label{eq_encoder_decoder_obj}
\min_\theta\sum_{i=1}^N\left\{-\log   P\left ( \mathbf{S}^{i} | \mathbf{V}^i; \theta \right ) \right\}.
\end{equation}

\subsection{Reconstructor}
As shown in Fig.~\ref{fig2}, the proposed reconstructor is built on the top of the encoder-decoder, which is expected to reproduce the video from the hidden state sequence of the decoder. However, due to the diversity and high dimension of the video frames, directly reconstructing the video frames seems to be intractable. Therefore, in this paper, the reconstructor aims at reproducing the sequential video frame representations generated by the encoder, with the hidden states $\mathbf{H}=\left \{ h_{1},h_{2},...,h_{n} \right \}$ of the decoder as input. The benefits of such a structure are two-fold. First, the proposed encoder-decoder-reconstructor architecture can be trained in an end-to-end fashion. Second, with such a reconstruction process, the decoder is encouraged to embed more information from the input video sequence. Therefore, the relationships between the video sequence and caption can be further enhanced, which is expected to improve the video captioning performance.
In practice, the reconstructor is realized by LSTMs. Two different architectures are customized to summarize the hidden states of the decoder for video feature reproduction. More specifically, one focuses on reproducing the global structure of the provided video, while the other pays more attentions to the local structure by selectively attending to the hidden state sequence.

\subsubsection{Reconstructing Global Structure}
\begin{figure}
  \centering
  \includegraphics[width=1\linewidth]{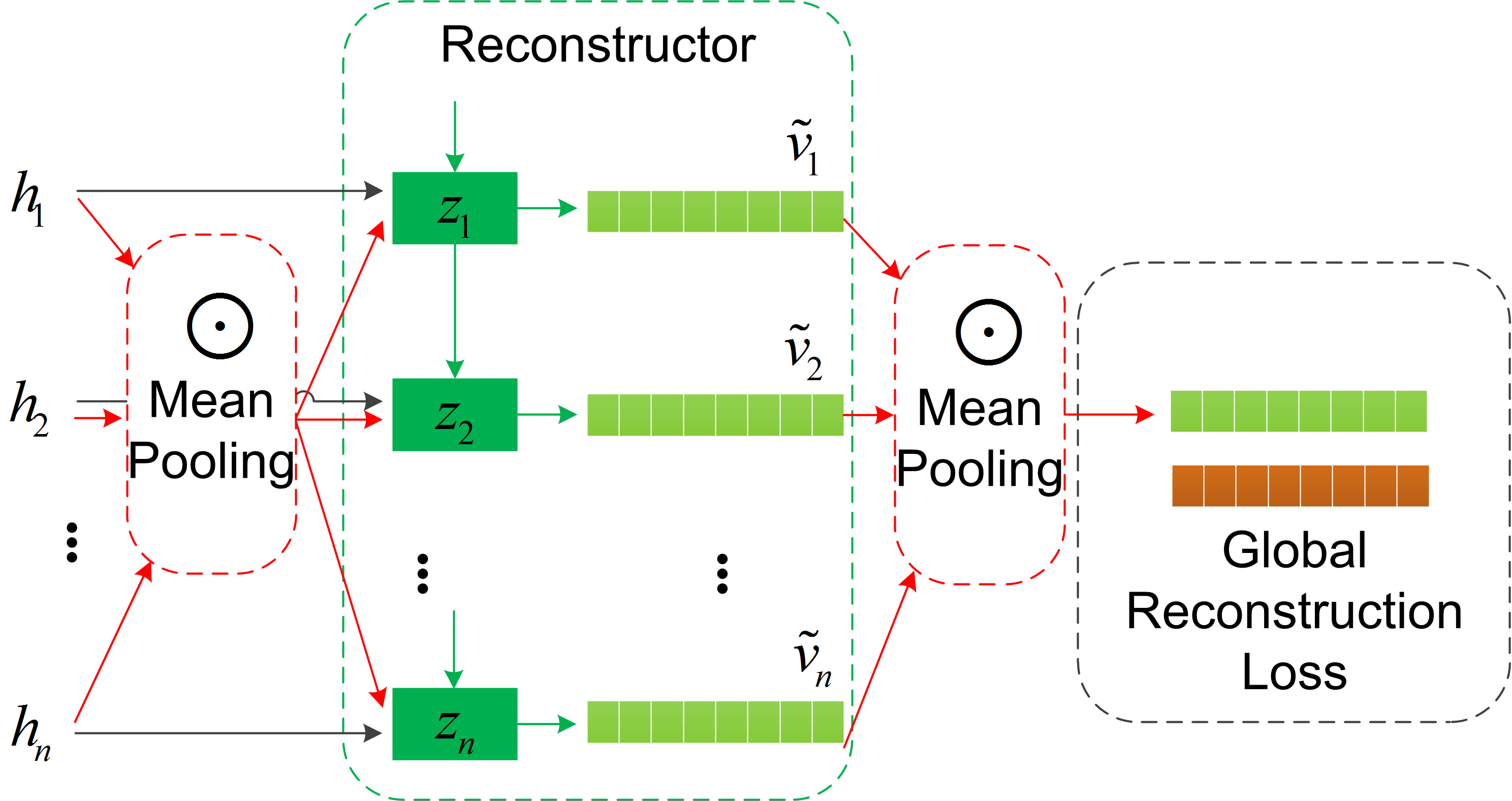} 
  \caption{An illustration of the proposed reconstructor that reproduces the global structure of the video sequence. {The left mean pooling is employed to summarize the hidden states of the decoder for the global representation of the caption. The reconstructor aims to reproduce the feature representation of the whole video by mean pooling (the right one) using the global representation of the caption as well as the hidden state sequence of the decoder.}}
  \label{fig3}
\end{figure}

The architecture for reconstructing the global structure of the video sequence is illustrated in Fig.~\ref{fig3}.
The whole sentence is fully considered to reconstruct the video global structure. Therefore, besides the hidden state $h_t$ at each time step, the global representation characterizing the semantics of the whole sentence is also taken as the input at each step.
Several methods like LSTM and multiple-layer perception, can be employed to fuse the hidden sequential states of the decoder to generate the global representation. Inspired by \cite{DBLP:conf/iccv/VenugopalanRDMD15}, the mean pooling strategy is performed on the hidden states of the decoder to yield the global representation of the caption:
\begin{equation}
\phi \left ( \mathbf{H} \right )=\frac{1}{n}\sum_{i=1}^{n}h_{i} ,
\end{equation}
where $\phi \left ( \cdot  \right )$ denotes the mean pooling process, which yields a vector representation $\phi \left ( \mathbf{H} \right )$ with the same size as $h_{i}$. Thus, the LSTM unit of the reconstructor is further modified as:
\begin{align}
\label{eq_lstm_mp}
\begin{split}
  \begin{pmatrix}i_t \\ f_t \\ o_t \\ g_t \end{pmatrix} &=
  \begin{pmatrix} \sigma \\ \sigma \\ \sigma \\ \tanh \end{pmatrix}
  \mathbf{T}
  \begin{pmatrix} h_t \\ z_{t-1}\\ \phi \left ( \mathbf{H} \right )\end{pmatrix}, \\
  m_t &= f_t \odot m_{t-1} + i_t \odot g_t, \\
  z_t &= o_t \odot \tanh(m_t),
\end{split}
\end{align}
where $i_t$, $f_t$, $m_t$, $o_t$, and $z_t$ denote the input, forget, memory, output, and hidden states of each LSTM unit, respectively. $\sigma$ and $\odot$ denote the logistic sigmoid activation and the element-wise multiplication, respectively.

To reconstruct the video global structure from the hidden state sequence produced by the encoder-decoder, the global reconstruction loss is defined as:
\begin{equation}
\label{eq_global_loss}
\mathcal{L}^g_{rec} = \psi\big(\phi(\mathbf{V}),\phi(\mathbf{Z})\big),
\end{equation}
where $\phi(\mathbf{V})$ denotes the mean pooling process on the video frame features, yielding the ground-truth global structure of the input video sequence. $\phi(\mathbf{Z})$ works on the hidden states of the reconstructor, indicating the global structure recovered from the captions. The reconstruction loss is measured by $\psi(\cdot)$, which is simply chosen as the Euclidean distance.

\subsubsection{Reconstructing Local Structure}
\label{sec:local_struct}
\begin{figure}
  \centering
  \includegraphics[width=1\hsize]{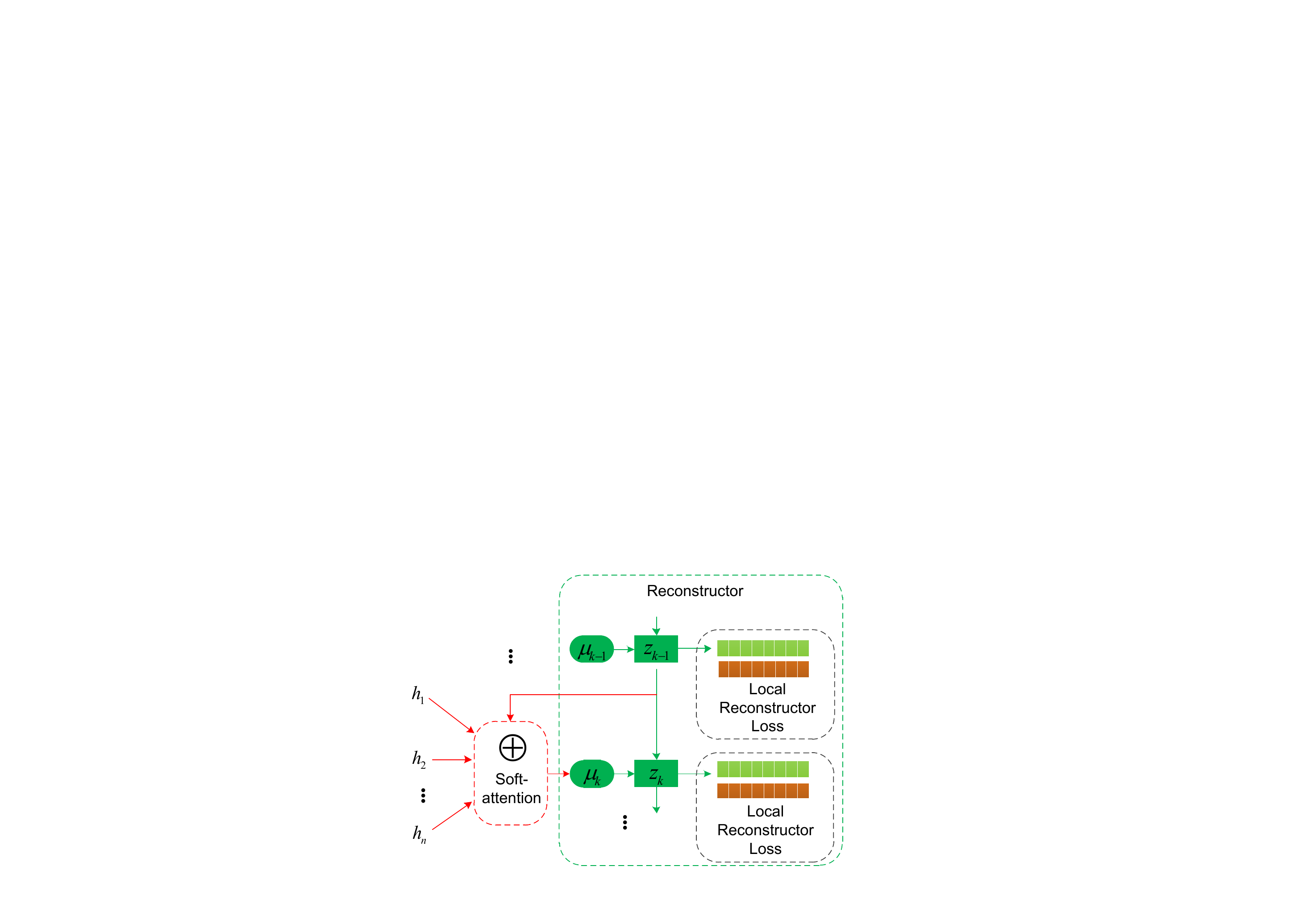} 
  \caption{An illustration of the proposed reconstructor that reproduces the local structure of the video sequence. The reconstructor works on the hidden states of the decoder by selectively adjusting the attention weight, and reproduces the feature representation frame by frame.}
  \label{fig4}
\end{figure}

The aforementioned reconstructor aims to reproduce the global representation for the whole video sequence, while neglects the local structures in each frame. In this subsection, we propose to learn and preserve the temporal dynamics by reconstructing each video frame as shown in Fig.~\ref{fig4}. Differing from the global structure estimation, we intend to reproduce the feature representation of each frame from the key hidden states of the decoder selected by the attention strategy \cite{DBLP:journals/corr/BahdanauCB14,DBLP:conf/iccv/YaoTCBPLC15}:

\begin{equation}
\label{eq:context}
\mu_t=\sum_{i=1}^{n}\beta_{i}^{t}h_{i},
\end{equation}
where $\sum_{i=1}^{n}\beta_i^t=1$ and $\beta_i^t$ denotes the weight computed for the $i_{th}$ hidden state at time step $t$ by the attention mechanism. Similar to Eq.~\ref{eq:3}, $\beta_i^t$ measures the relevance of the $i_{th}$ hidden state in the caption given all the previously reconstructed frame representations $\{z_1,z_2,\dots,z_{t-1}\}$. Such a strategy encourages the reconstructor to work on the hidden states selectively by adjusting the attention weight $\beta_i^t$ and yield the context information $\mu_t$ at each time step as in Eq.~\ref{eq:context}. As such, the proposed reconstructor can further exploit the temporal dynamics and the word compositions across the whole caption. The LSTM unit is thereby reformulated as:

\begin{align}
\label{eq_lstm_sa}
\begin{split}
  \begin{pmatrix}i_t \\ f_t \\ o_t \\ g_t \end{pmatrix} &=
  \begin{pmatrix} \sigma \\ \sigma \\ \sigma \\ \tanh \end{pmatrix}
  \mathbf{T}
  \begin{pmatrix} \mu_t \\ z_{t-1} \end{pmatrix}.
\end{split}
\end{align}

Differing from the global structure recovery step in Eq.~\ref{eq_lstm_mp}, the dynamically generated context $\mu_t$ is taken as the input other than the hidden state $h_t$ and its mean pooling representation $\phi \left ( \mathbf{H} \right )$. Moreover, instead of directly generating the global mean representation of the whole video sequence, we propose to produce the feature representation frame by frame. 
The reconstruction loss is thereby defined as:
\begin{equation}
\label{eq_local_loss}
\mathcal{L}^l_{rec} = \frac{1}{m}\sum_{j=1}^{m}\psi(z_{j},\mathbf{v}_{j}).
\end{equation}


\subsection{Training}
Formally, we train the proposed encoder-decoder-reconstructor architecture by minimizing the whole loss defined in Eq.~\ref{eq_full_loss}, which involves both the forward (video-to-sentence) likelihood and the backward (sentence-to-video) reconstruction loss:
\begin{equation}
\label{eq_full_loss}
\begin{split}
\mathcal{L}(\theta,\theta_{rec}) = \sum_{i=1}^N & \Big( \underbrace{-\log P\left ( \mathbf{S}^{i} | \mathbf{V}^i; \theta \right )}_{\text{encoder-decoder}}\\
&+ \lambda \underbrace{\mathcal{L}_{rec}(\mathbf{V}^i,\mathbf{Z}^i;\theta_{rec})}_{\text{reconstructor}}   \Big).
\end{split}
\end{equation}

The reconstruction loss $\mathcal{L}_{rec}(\mathbf{V}^i,\mathbf{Z}^i;\theta_{rec})$ can be realized by the global loss in Eq.~\ref{eq_global_loss} or the local loss in Eq.~\ref{eq_local_loss}. The hyper-parameter $\lambda$ is introduced to seek a trade-off between the encoder-decoder and the reconstructor.

The training of our proposed RecNet model proceeds in two stages. First, we rely on the forward likelihood 
to train the encoder-decoder component of the RecNet, which is terminated by the early stopping strategy. Afterwards, the reconstructor and the backward reconstruction loss $\mathcal{L}_{rec}(\theta_{rec})$ are introduced. We use the whole loss defined in Eq.~\ref{eq_full_loss} to jointly train the reconstructor and fine-tune the encoder-decoder. For the reconstructor, the reconstruction loss is calculated using the hidden state sequence generated by the LSTM units in the reconstructor as well as the video frame feature sequence.

\section{Experimental Results}
In this section, we evaluate the proposed video captioning method on benchmark datasets such as Microsoft Research video to text (MSR-VTT) \cite{xu2016msr} dataset and Microsoft Research Video Description Corpus (MSVD) \cite{chen2011collecting}. To compare with existing work, we compute the popular metrics including BLEU-4 \cite{papineni2002bleu}, METEOR \cite{banerjee2005meteor}, ROUGE-L \cite{lin2004rouge}, and CIDEr \cite{vedantam2015cider} with the codes released on the Microsoft COCO evaluation server \cite{chen2015microsoft}.

\subsection{Datasets and Implementation Details}
\textbf{MSR-VTT.} It is the largest dataset for video captioning so far in terms of the number of video-sentence pairs and the vocabulary size. In the experiments, we use the initial version of MSR-VTT, referred as MSR-VTT-10K, which contains 10K video clips from 20 categories. Each video clip is annotated with 20 sentences by 1327 workers from Amazon Mechanical Turk. Therefore, the dataset results in a total of 200K clip-sentence pairs and 29,316 unique words.
We use the public splits for training and testing, \textit{i.e.}, 6513 for training, 497 for validation, and 2990 for testing.

\textbf{MSVD.} It contains 1970 YouTube short video clips, and each one depicts a single activity in 10 seconds to 25 seconds.
and each video clip has roughly 40 English descriptions. Similar to the prior work \cite{pan2016jointly,DBLP:conf/iccv/YaoTCBPLC15}, we take 1200 video clips for training, 100 clips for validation and 670 clips for testing.

For the sentences, we remove the punctuations, split them with blank space and convert all words into lowercase.
{The sentences longer than 30 are truncated, and the word embedding size for each word is set to 468.} 

For the encoder, we feed all frames of each video clip into Inception-V4 \cite{DBLP:conf/aaai/SzegedyIVA17} which is pretrained on the ILSVRC-2012-CLS \cite{russakovsky2015imagenet} classification dataset for feature extraction after resizing them to the standard size of $299\times 299$, and extract the 1536 dimensional semantic feature of each frame from the last pooling layer.
Inspired by \cite{DBLP:conf/iccv/YaoTCBPLC15}, we choose the equally-spaced 28 features from one video, and pad them with zero vectors if the number of features is less than 28. The input dimension of the decoder is 468, the same to that of the word embedding, while the hidden layer contains 512 units. For the reconstructor, the inputs are the hidden states of the decoder and thus have the dimension of 512. To ease the reconstruction loss computation, the dimension of the hidden layer is set to 1536 same to that of the frame features produced by the encoder.

During the training, the AdaDelta \cite{zeiler2012adadelta} is employed for optimization. The training stops when the CIDEr value on the validation dataset stops increasing in the following 20 successive epochs. In the testing, beam search with size 5 is used for the final caption generation.

\subsection{Study on the Encoder-Decoder}
\begin{table}[!t]
\renewcommand\arraystretch{1.5}
  \caption{Performance evaluation of different video captioning models on the testing set of the MSR-VTT dataset
  (\%). The encoder-decoder framework is equipped with different CNN structures such as AlexNet, GoogleNet, VGG19 and Inception-V4. Except Inception-V4, the metric values of the other published models are referred from the work in \cite{msr-vtt-large-video-description-dataset-bridging-video-language-supplementary-material}, and the symbol ``-'' indicates that such metric is unreported.}
  \label{table1}
\scriptsize
  \centering
  \vspace{15pt}
  \begin{tabular}{c|c|c|c|c}
  \hline
  Model                             & BLEU-4  &  METEOR  &  ROUGE-L  &  CIDEr \\ \hline \hline
  MP-LSTM (AlexNet)                 & 32.3    &  23.4    & -         &  -     \\
  MP-LSTM (GoogleNet)               & 34.6    &  24.6    &  -        &  -     \\
  MP-LSMT (VGG19)                   & 34.8    &  24.7    &  -        &  -     \\
  SA-LSTM (AlexNet)                 & 34.8    &  23.8    &  -        &  -     \\
  SA-LSTM (GoogleNet)               &  35.2   &  25.2    &  -        &  -     \\
  SA-LSTM (VGG19)                   & 35.6    &  25.4    &  -        &  -     \\
  SA-LSTM (Inception-V4)            & 36.3    &  25.5    &  58.3     &  39.9  \\  \hline
  RecNet$_{global}$                 & 38.3    &  26.2    &  59.1     &  41.7  \\
  RecNet$_{local}$          & \textbf{39.1}    &  \textbf{26.6}    &  \textbf{59.3}     &  \textbf{42.7}  \\  \hline
  \end{tabular}
\end{table}
In this section, we first test the impacts of different encoder-decoder architectures in video captioning, such as SA-LSTM and MP-LSTM. Both are popular encoder-decoder models and share similar LSTM structure, except that SA-LSTM introduced an attention mechanism to aggregate frame features, while MP-LSTM relies on the mean pooling. As shown in Table~\ref{table1}, with the same encoder VGG19, SA-LSTM yielded 35.6 and 25.4 on the BLEU-4 and METEOR respectively, while MP-LSTM only produced 34.8 and 24.7, respectively. The same results can be obtained when using AlexNet and GoogleNet as the encoder. Hence, it is concluded that exploiting the temporal dynamics among frames with attention mechanism performed better in sentence generation than mean pooling on the whole video.

\begin{figure} [!htbp]
  \centering
  \vspace{7.5pt}
  \includegraphics[width=\hsize]{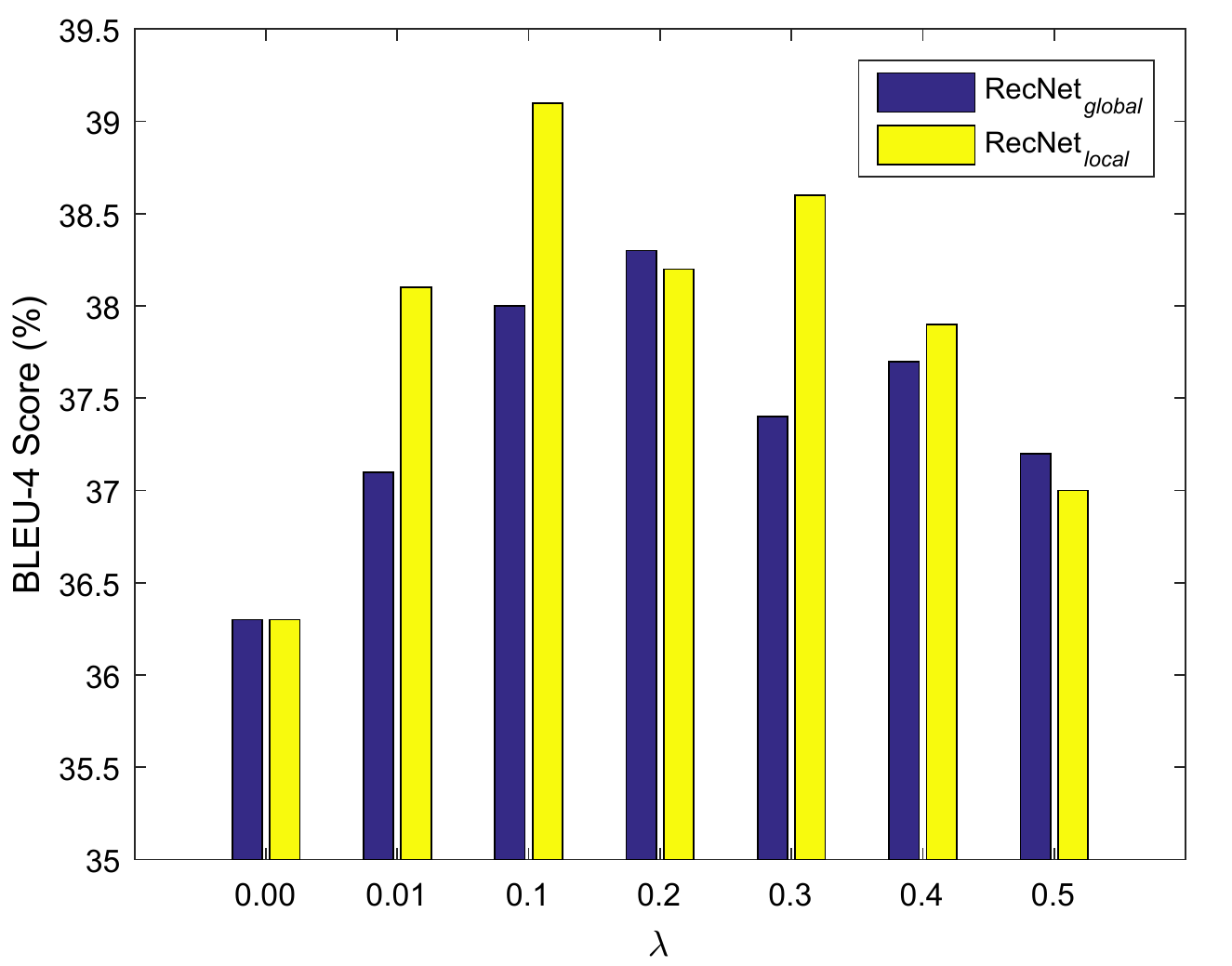} 
  \caption{Effects of the trade-off parameter $\lambda$ for RecNet$_{global}$ and RecNet$_{local}$ in terms of BLEU-4 metric on MSR-VTT. It is noted that $\lambda=0$ means the reconstructor is off, and the RecNet turns to be a conventional encoder-decoder model.}
  \label{global_local_curve}
  \vspace{7.5pt}
\end{figure}

\begin{figure*}[htbp]
  \centering
  \vspace{7.5pt}
  \includegraphics[width=0.97\hsize]{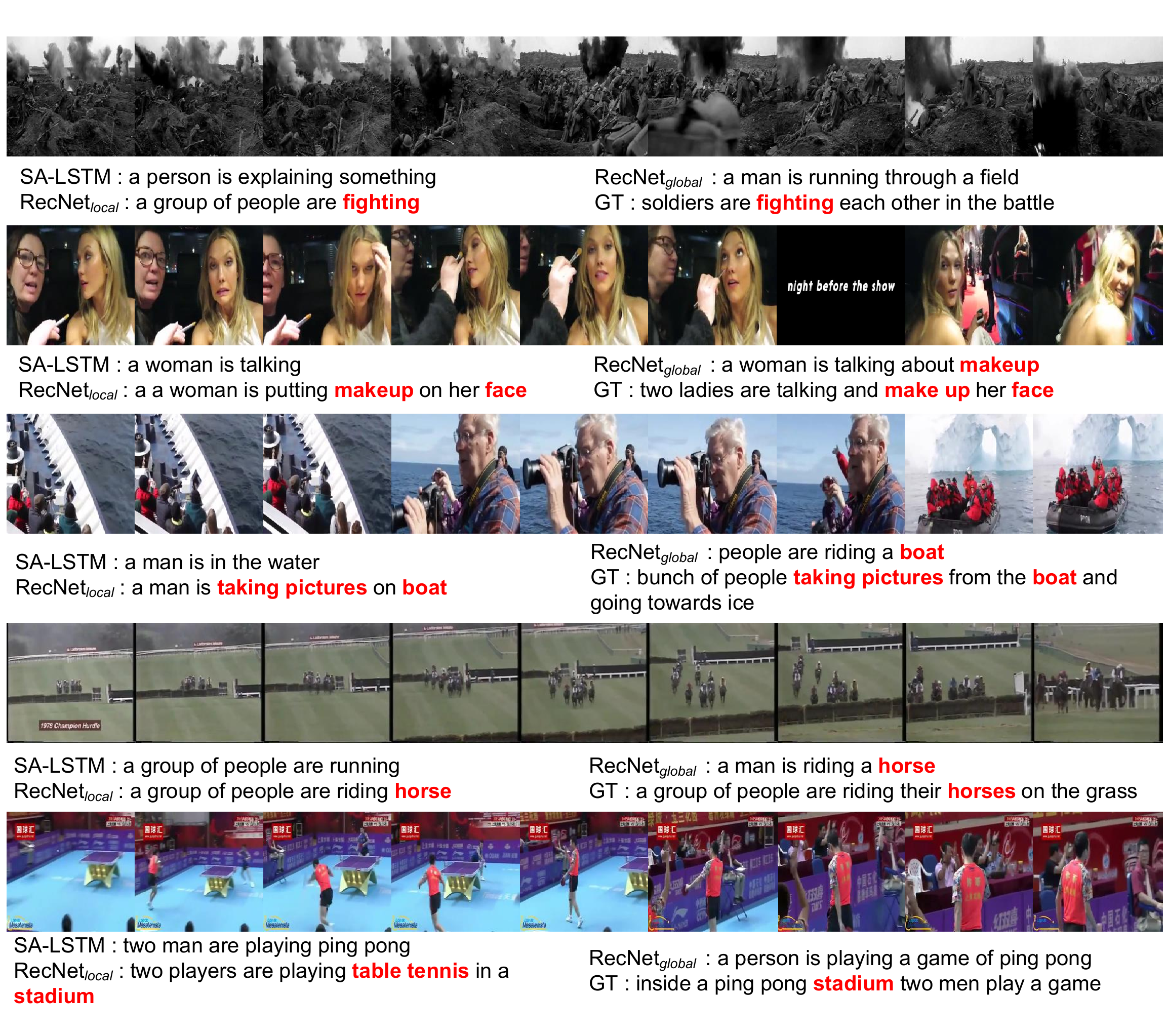} 
  \caption{Visualization of some video captioning examples on the MSR-VTT dataset with different models. Due to the page limit, only one ground-truth sentence is given as reference. Compared to SA-LSTM, the proposed RecNet is able to yield more vivid and descriptive words highlighted in red boldface, such as ``\texttt{fighting}'', ``\texttt{makeup}'', ``\texttt{face}'', and ``\texttt{horse}''.}
  \label{samples}
  \vspace{7.5pt}
\end{figure*}
Besides, we also introduced Inception-V4 as an alternative CNN for feature extraction in the encoder. It is observed that with the same encoder-decoder structure SA-LSTM, Inception-V4 yielded the best captioning performance comparing to the other CNNs such as AlexNet, GoogleNet, and VGG19. This is probably because Inception-V4 is a deeper network and better at semantic feature extraction. Hence, SA-LSTM equipped with Inception-V4 is employed as the encoder-decoder model in the proposed RecNet.

By adding the global or local reconstructor to the encoder-decoder model SA-LSTM, we can have the proposed encoder-decoder-reconstructor architecture: RecNets. Apparently, such structure provided significant gains to the captioning performance in all metrics. This proved the backward flow information introduced by the proposed reconstructor could encourage the decoder to embed more semantic information and also regularize the generated caption to be more consistent with the video contents. More discussion about the proposed reconstrucutor will be given in Section \ref{sec:recon}.

\subsection{Study on the Trade-off Parameter $\lambda$}
In this section, we discuss the influence of the trade-off parameter $\lambda$ in Eq.~\ref{eq_full_loss}. With different $\lambda$ values, the obtained BLEU-4 metric values are given in Figure \ref{global_local_curve}. First, it can be concluded again that adding the reconstruction loss ($\lambda>0$) did improve the performance of video captioning in terms of BLEU-4. Second, there is a trade-off between the forward likelihood loss and the backward reconstruction loss, as too large $\lambda$ may incur noticeable deterioration in caption performance. Thus, $\lambda$ needs to be more carefully selected to balance the contributions of the encoder-decoder and the reconstructor. As shown in Figure \ref{global_local_curve}, we empirically set $\lambda$ to 0.2 and 0.1 for RecNet$_{global}$ and RecNet$_{local}$, respectively.

\subsection{Study on the Reconstructors}
\label{sec:recon}
The difference of the proposed two reconstructors is discussed in this section. The quantitative results of RecNet$_{local}$ and RecNet$_{global}$ on MSR-VTT are given on the bottom two rows of Table~\ref{table1}. It can be observed that RecNet$_{local}$ performs slightly better than RecNet$_{global}$. The reason mainly lies in the temporal dynamic modeling. RecNet$_{global}$ employs mean pooling to reproduce the video representation and misses the local temporal dynamics, while the attention mechanism is included in RecNet$_{local}$ to exploit the local temporal dynamics for each frame reconstruction.

However, the performance gap between RecNet$_{global}$ and RecNet$_{local}$ is not significant. One possible reason is that the visual information of frames is very similar. As the video clips of MSR-VTT are short, the visual representations of frames have few differences with each other, that is the global and local structure information is similar.
Another possible reason is the complicated video-sentence relationship, which may lead to similar input information for RecNet$_{global}$ and RecNet$_{local}$.

\subsection{Qualitative Analysis}

Besides, some qualitative examples are shown in Fig.~\ref{samples}. Still, it can be observed that the proposed RecNets with local and global reconstructors generally produced more accurate captions than the typical encoder-decoder model SA-LSTM. For example, in the second example, SA-LSTM generated ``\texttt{a woman is talking}'', which missed the core subject of the video, \textit{i.e.}, ``\texttt{makeup}''. By contrast, the captions produced by RecNet$_{global}$ and RecNet$_{local}$ are ``\texttt{a woman is talking about makeup}'' and  ``\texttt{a women is putting makeup on her face}'', which apparently are more accurate. RecNet$_{local}$ even generated the word of ``\texttt{face}'' which results in a more descriptive caption. More results can be found in the supplementary file.

\begin{table}[!h]
\renewcommand\arraystretch{1.5}
\caption{Performance evaluation of different video captioning models on the MSVD dataset
in terms of BLEU-4, METEOR, ROUGE-L, and CIDEr scores (\%). The symbol "-" indicates such metric is unreported.
}
  \label{table2}
  \scriptsize
  \centering
  \vspace{15pt}
  \begin{tabular}{c|c|c|c|c}
  \hline
  Model                                              & BLEU-4  &  METEOR  &  ROUGE-L  &  CIDEr \\ \hline \hline
  MP-LSTM (AlexNet)\cite{venugopalan2015translating} & 33.3    &  29.1    & -         &  -     \\
  GRU-RCN\cite{ballas2015delving}					 & 43.3	   &  31.6    &  -		  &  68.0  \\
  HRNE\cite{pan2016hierarchical}					 & 43.8    &  33.1    &  -   	  &  -     \\
  LSTM-E\cite{pan2016jointly}                        & 45.3    &  31.0    &  -		  &  -     \\
  LSTM-LS (VGG19)\cite{liu2017video}                 & 46.5    &  31.2    &  -        &  -     \\
  h-RNN\cite{yu2016video}							 & 49.9    &  32.6    &  -        &  65.8  \\
  aLSTMs \cite{gao2017video}                         & 50.8    &  33.3    & -         &  74.8  \\  \hline
  S2VT (Inception-V4)                                & 39.6    &  31.2    &  67.5     &  66.7  \\
  SA-LSTM (Inception-V4)                             & 45.3    &  31.9    &  64.2     &  76.2  \\   \hline
  RecNet$_{global}$ (S2VT)                           & 42.9    &  32.3    &  68.5     &  69.3  \\
  RecNet$_{local}$  (S2VT)                           & 43.7    &  32.7    &  68.6     &  69.8  \\
  RecNet$_{global}$ (SA-LSTM)                        & 51.1    &  34.0    &  69.4     &  79.7  \\
  RecNet$_{local}$  (SA-LSTM)                 &\textbf{52.3} &\textbf{34.1}  &\textbf{69.8} &  \textbf{80.3}  \\   \hline
  \end{tabular}
\end{table}

\subsection{Evaluation on the MSVD Dataset}
\label{sec:msvd}
Finally, we tested the proposed RecNet on the MSVD dataset \cite{chen2011collecting}, and compared it to more benchmark encoder-decoder models, such as GRU-RCN\cite{ballas2015delving}, HRNE\cite{pan2016hierarchical}, h-RNN\cite{yu2016video}, LSTM-E\cite{pan2016jointly}, aLSTMs\cite{gao2017video} and LSTM-LS\cite{liu2017video}. The quantitative results are given in Table~\ref{table2}. It is observed that the RecNet$_{local}$ and RecNet$_{global}$ with SA-LSTM performed the best and second best in all metrics, respectively. Besides, we introduced the reconstructor to S2VT\cite{DBLP:conf/iccv/VenugopalanRDMD15} to build another encoder-decoder-reconstructor model. The results show that both global and local reconstructors bring improvement to the original S2VT in all metrics, which again demonstrate the benefits of video captioning based on bidirectional cue modeling.


\section{Conclusions}
In this paper, we proposed a novel RecNet with the encoder-decoder-reconstructor architecture for video captioning, which exploits the bidirectional cues between natural language description and video content. Specifically, to address the backward information from description to video, two types of reconstructors were devised to reproduce the global and local structures of the input video, respectively. The forward likelihood and backward reconstruction losses were jointly modeled to train the proposed network. The experimental results on  the benchmark datasets corroborate the superiority of the proposed RecNet over the existing encoder-decoder models in video caption accuracy.

\section*{Acknowledgments}
The authors would like to thank the anonymous reviewers for the constructive comments to improve the paper. This work was supported by the NSFC Grant no. 61573222, Shenzhen Future Industry Special Fund JCYJ20160331174228600, Major Research Program of Shandong Province 2015ZDXX0801A02, National Key Research and Development Plan of China under Grant 2017YFB1300205 and Fundamental Research Funds of Shandong University 2016JC014.


{\small
\bibliographystyle{ieee}
\bibliography{egbib}
}

\end{document}